\documentclass{article}

\usepackage{amsmath}
\usepackage{subfigure}
\usepackage[svgnames]{xcolor} 
\usepackage{graphicx}
\usepackage[margin= 0.8in]{geometry}
\usepackage[toc,page]{appendix}

\def\eg{\emph{e.g}}

\def\etal{\emph{et al}}
\def\ie{\emph{i.e}}

\newcommand*{\titleAT}{\begingroup 
\newlength{\drop} 
\drop=0.05\textheight 

\includegraphics[scale=1.5]{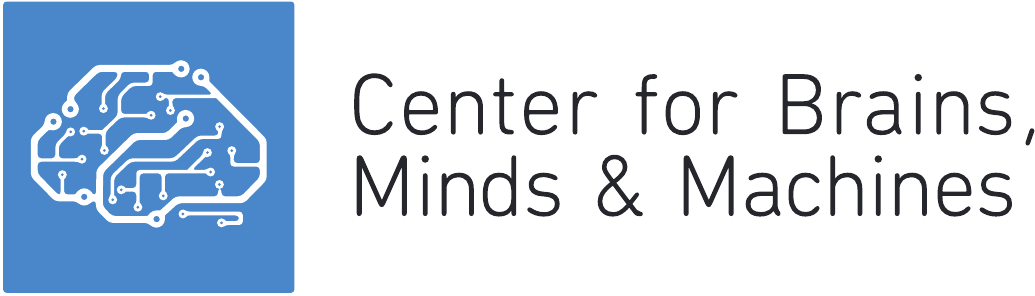}

\textcolor{CornflowerBlue}{\rule{\textwidth}{3 pt}}\par 
\vspace{2pt}\vspace{-\baselineskip} 
\rule{\textwidth}{0.4pt}\par 

\vspace{\drop} 
\textbf{\textsf{\large{CBMM Memo No. \memonumber}}}\quad \quad \quad\quad \quad \quad \quad\quad\quad \quad\quad\quad      \textbf{\large{\memodate}}

\vspace{\drop}
\begin{center}
\textbf{\textsf{\huge{\memotitle}}}\\
\vspace{0.4\drop}
\textbf{\Large{\textsf{by}}}\\
\vspace{0.4\drop}
\textbf{\textsf{\large{\memoauthors}}}
\end{center}
\vspace{\drop}
\textbf{\textsf{\large{\noindent Abstract}:}} {\memoabstract}

\textcolor{CornflowerBlue}{\rule{\textwidth}{3 pt}}\par 
\vspace{2pt}\vspace{-\baselineskip} 
\rule{\textwidth}{0.4pt}\par

\begin{minipage}{.15\linewidth}
\includegraphics[scale=0.1]{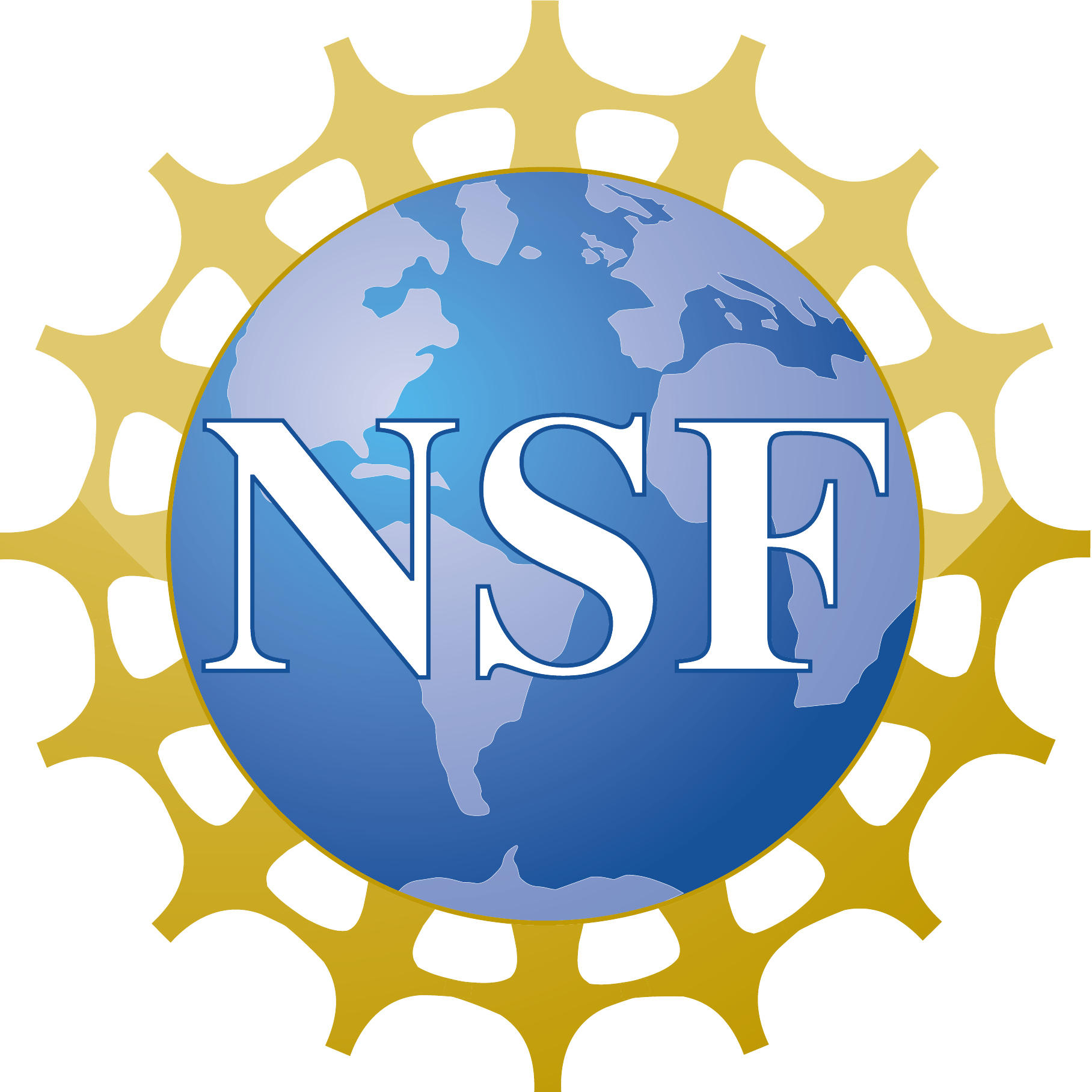}
\end{minipage}
\begin{minipage}{.84\linewidth}
\textbf{\textsf{\large{This work was supported by the Center for Brains, Minds and Machines (CBMM), funded by NSF STC award  CCF - 1231216.}}}
\end{minipage}
\endgroup}

\begin{document}

\pagestyle{empty} 

\def\eg{\textsf{\it e.g.}}
\def\ie{\textsf{\it i.e.}}

\def\memonumber{ \textsf{018}} 
\def\memodate{\textsf{\today}} 
\def\memotitle{\textsf{Parsing Semantic Parts of Cars Using Graphical Models and Segment Appearance Consistency}} 
\def\memoauthors{\textsf{Wenhao Lu$^{1}$,Xiaochen Lian$^{2}$,Alan Yuille$^{2}$ }\\
$^{1}$Tsinghua Univesity~~
$^{2}$University of California, Los Angeles~~\\
{\tt\small yourslewis@gmail.com~lianxiaochen@ucla.edu~yuille@stat.ucla.edu}}

\def\memoabstract{\textsf{This paper addresses the problem of semantic part parsing (segmentation) of cars, \ie assigning every pixel within the car to one of the parts (\eg body, window, lights, license plates and wheels). We formulate this as a landmark identification problem, where a set of landmarks specifies the boundaries of the parts. A novel mixture of graphical models is proposed, which dynamically couples the landmarks to a hierarchy of segments. When modeling pairwise relation between landmarks, this coupling enables our model to exploit the local image contents in addition to spatial deformation, an aspect that most existing graphical models ignore. In particular, our model enforces appearance consistency between segments within the same part. Parsing the car, including finding the optimal coupling between landmarks and segments in the hierarchy, is performed by dynamic programming. We evaluate our method on a subset of PASCAL VOC 2010 car images and on the car subset of 3D Object Category dataset (CAR3D). We show good results and, in particular, quantify the effectiveness of using the segment appearance consistency in terms of accuracy of part localization and segmentation.}}

\titleAT 
		
\newpage
\section{Introduction}
\label{sec:intro}
\begin{figure}[t]
\centering
\includegraphics[width=0.9\textwidth]{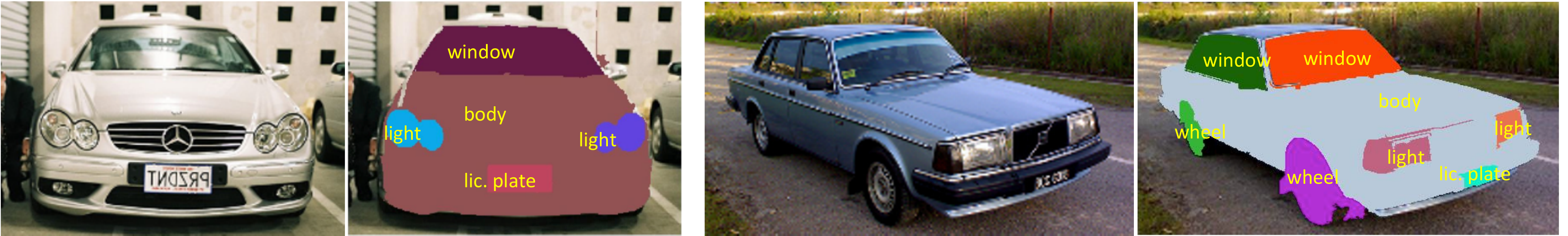}
\caption{The goal of car parsing is to detect the locations of semantic parts and to perform object part segmentation. The inputs (left) are images of a car taken from different viewpoints. The outputs (right) are the locations of the car parts -- the wheels, lights, windows, license plates and bodies -- so that each pixel within the car is assigned to a part.}
\label{fig:goal}
\end{figure}

This paper addresses the two goals of parsing an object into its semantic parts and performing object part segmentation, so that each pixel within the object is assigned to one of the parts (i.e. all pixels in the object are labeled). More specifically, we attempt to parse cars into wheels, lights, windows, license plates and body, as illustrated in Figure \ref{fig:goal}. This is a fine-scale task, which differs from the classic task of detecting an object by estimating a bounding box. 

\begin{figure*}[!t]
\centering
\includegraphics[width=1.0\textwidth]{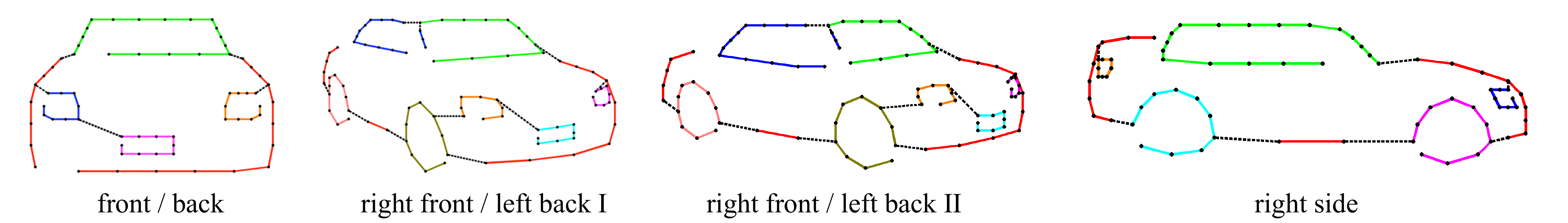}
\caption{The proposed mixture-of-trees model. Models of left-front, right-back and right views are not shown due to the symmetry. The landmarks connected by the solid lines of same colors belong to the same semantic parts. The black dashed lines show the links between different parts. Best view in color.}\label{fig:tree_model}
\end{figure*}

We formulate the problem as landmark identification. We first select representative locations on the boundaries of the parts to serve as landmarks. They are selected so that locating them yields the silhouette of the parts, and hence enables us to do object part segmentation. We use a mixture of graphical models to deal with different viewpoints so that we can take into account how the visibility and appearance of parts alter with viewpoint (see Figure \ref{fig:tree_model}). 

\begin{figure}[t]
\centering
\includegraphics[width=0.8\textwidth]{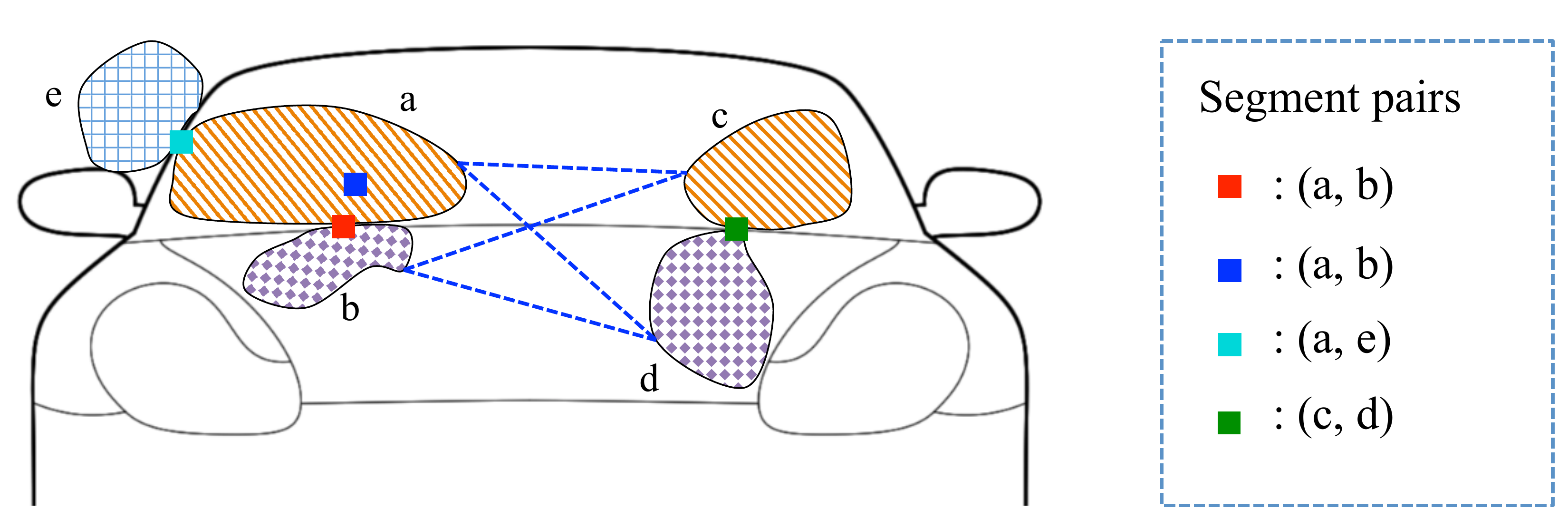}
\caption{Illustration of segmentation appearance consistency (SAC) and segment pairs. Red and green squares represent two neighboring landmarks lying on the boundary between window and body. Each landmark has two segments ($a$ and $b$ for the red landmark, $c$ $d$ for the green landmark) close to it. Our method models and learns the SACs for every pair of neighboring landmarks (blue dashed lines) and uses them to enhance the reliability of landmark localization. For the blue landmark, its segment pair is the same as the red landmark, which is the closest one on the boundary.}
\label{fig:motivation}
\end{figure}

A novel aspect of our graphical model is that we couple the landmarks with the segmentation of the image to exploit the image contents when modeling the pairwise relation between neighboring landmarks. In the ideal case where part boundaries of car are all preserved by the segmentation, we can assume that the landmarks lie near the boundaries between different segments. Each landmark is then associated to the appearance of its two closest segments. This enables us to associate appearance information to the landmarks and to introduce pairwise coupling terms which enforce that the appearance is similar within parts and different between parts. We call this segmentation appearance consistency (SAC) between segments of neighboring landmarks. This is illustrated in Figure \ref{fig:motivation}, where Both of the two neighboring landmarks (the red and green squares) on the boundary between the window and the body have two segments (belonging to window and body respectively) close to them. Segments from the same part tend to have homogeneous color and texture appearance (\eg $a$ and $c$, $b$ and $d$ in the figure), while segments from different parts usually do not (\eg $a$ and $b$, $c$ and $d$ in the figure). The four blue dashed lines in the figure correspond to the SAC terms whose strengths will be learnt. 

However, in practice, it is always impossible to capture all part boundaries using single level segmentation. Instead, people try to use a pool of segmentations \cite{gould2009decomposing,chen2011piecing,kumar2010efficiently} or segmentation trees \cite{2011_pami_Arbelaez,lempitsky2011pylon,veksler2000image}. Inspired by those, we couple the landmarks to a hierarchical segmentation of the image. However, the difference of the sizes of the parts (\eg the license plate is much smaller than the body) and the variability of the images mean that the optimal segmentation level for each part also varies. Therefore the level of the hierarchy used in this coupling must be chosen {\it dynamically} during inference/parsing. This leads us to treat the level of the hierarchy for each part as a hidden variable. By doing this, our model is able to automatically select the most suitable segmentation level for each part while parsing the image.

\section{Related Work\label{sec:related}}

There is an extensive literature dating back to Fischler and Ershlager \cite{fischler1973representation} which represents objects using graphical models. Nodes of the graphs typically represent distinctive regions or landmark points. These models are typically used for detecting objects \cite{2008_cvpr_felzenszwalb,2010_pami_felzenszwalb} but they can also be used for parsing objects by using the positions of the nodes to specify the locations of different parts of the object. For example, Zhu \etal \cite{zhu2008max,zhu2010learning} uses an compositional AND/OR graph to parse baseball players and horses. More recently, in Zhu and Ramanan's graphical model for faces \cite{2012_cvpr_zhu} there are nodes which correspond to the eyes and mouth of the face. But we note that these types of models typically only output a parse of the object and are not designed to perform object part segmentation. They do not exploit the SAC either.

Recently, a very similar graphical model for cars has been proposed by Hejrati and Ramanan \cite{2012_nips_hejrati}, which cannot do part segmentation since each part is represented by only one node. The more significant difference is that the binary terms do not consider the local image contents.

There are, however, some recent graphical models that can perform object part segmentation. Bo and Fowlkes \cite{2011_cvpr_yhb} use a compositional model to parse pedestrians, where the semantic parts of pedestrians are composed of segments generated by the UCM algorithm \cite{2011_pami_Arbelaez} (they select high scoring segments to form semantic parts and use heuristic rules for pruning the space of parses). Thomas \etal \cite{2008_rss_thomas} use Implicit Shape Models to determine the semantic part label of every pixel. Eslami and Williams \cite{2012_nips_eslami} extend the Shape Bolzmann Machine to model semantic parts and enable object part segmentation. \cite{2008_rss_thomas} and \cite{2012_nips_eslami} did car part segmentation on ETHZ car dataset~\cite{2008_rss_thomas}, which contains non-occluded cars of a single view (semi-profile view).

Image labeling is a related problem since it requires assigning labels to pixels, such as \cite{shotton2009textonboost,2010_eccv_munoz,2011_pami_liu,2012_cvpr_eigen,farabet2012scene,tighe2013finding}. But these methods are applied to labeling all the pixels of an image, and are not intended to detect the position of objects or perform object part segmentation.

\section{The Method for Parsing Cars\label{sec:model}}

We represent the car and its semantic parts by a mixture of tree-structured graphical models, one model for each viewpoint. The model is represented by $\mathcal{G}=(\mathcal{V},\mathcal{E})$. The nodes $\mathcal{V}$ correspond to landmark points. They are divided into subsets $\mathcal{V} = \bigcup _{p=1}^N \mathcal{V}_p$, where $N$ is the number of parts and $\mathcal{V}_p$ consists of landmarks lying at the boundaries of semantic part $p$. The edge structures $\mathcal{E}$ are manually designed (see Figure \ref{fig:tree_model}).

We define an energy function for each graphical model, which consists of unary terms at the landmarks and binary terms at the edges. The binary terms not only model the spatial deformations as in\cite{2012_cvpr_zhu,2010_pami_felzenszwalb}, but also utilize local image contents, \ie the segment appearance consistency (SAC) between neighboring landmarks. 

\begin{figure}[!t]
\centering
\includegraphics[width=1.0\textwidth]{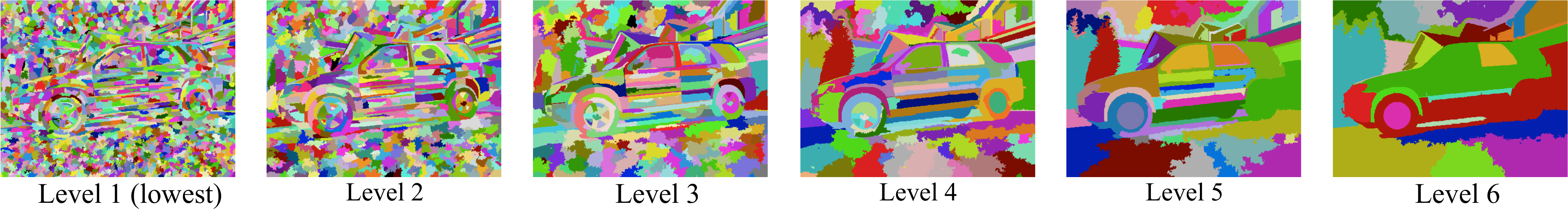}
\caption{The segments output by SWA at six levels. Note how the segments covering the semantic parts change from level 1 to level 6 (\eg left windows and left wheels). This illustrates that different parts need different levels of segmentation. For example, the best level for the left-back wheel is level 4 and the best level for the left windows is level 5. Best view in color.}\label{fig:swa_a}
\end{figure}

To do that, we couple the landmarks to a hierarchical segmentation of the image which is obtained by the SWA algorithm \cite{sharon06Hierarchy} (see Figure \ref{fig:swa_a} for a typical SWA segmentation hierarchy). Then we associate with each image location at every segmentation level a pair of nearby segments: If a location is on the segment boundary, the two segments are on either sides of the boundary, otherwise it shares the same segment pairs with the nearest boundary location. Then SAC terms are used to model the four pairings of segments from neighboring landmarks (blue dashed lines in Figure \ref{fig:motivation} for example). The strengths of SAC terms are learnt from data. In order to do the learning, the four pairing need to be ordered, or equivalently, the two segments of each location need to be represented in the form of an ordered tuple $(s^1,s^2)$. In practice, choosing two segments for a segment boundary location and ordering them is not straightforward (\eg a location on T-junction where there are more than two segments nearby). We put technical details about segment pairs in Section \ref{sec:impl_detail}.

\subsection{Score Function\label{sec:score_func}}

In this section we describe the score function for each graphical model, which is the sum of unary potentials defined at the graph nodes, representing the landmarks, and binary potentials defined over the edges connecting neighboring landmarks.

We first define the variables of the graph. Each node has pixel position of landmark $l_i = (x_i,y_i)$. The set of all positions is denoted by ${\bf L}=\{l_i\}_{i=1}^{|\mathcal{V}|}$. We denote by $p_i$ the indicator specifying which part landmark $i$ belongs to, and by $h(p)$ the segmentation level of part $p$. Then the segment pair of node $i$, $\bf{s}_i$, can be seen as the function of $h(p_i)$, which we denote by ${\bf s}_{i,h}$ for simplicity. Similar to the definitions of $\bf{L}$, we have ${\bf H}=\{h(p_i)\}_{i=1}^N$ and ${\bf S}({\bf H})=\{{\bf s}_{i,h}\}_{i=1}^{|\mathcal{V}|}$. The score function of the model for viewpoint $v$ is
\begin{equation}
S({\bf L},{\bf H},v \mid {\bf I}) = \phi({\bf L},{\bf H},v \mid {\bf I}) + \psi({\bf L},{\bf H},v \mid {\bf I}) + \beta_v \label{eq:energy}
\end{equation}
In the following we omit $v$ for simplicity. The unary terms $\phi({\bf L},{\bf H} \mid {\bf I})$ is expressed as:
\begin{equation}
\phi({\bf L},{\bf H} \mid {\bf I}) = \sum _{i \in \mathcal{V}} \left[{\bf w}_i^f \cdot f(l_i \mid {\bf I}) + w_i^e e(h(p_i),l_i \mid {\bf I})\right] \label{eq:unary}
\end{equation}
The first term in the bracket of Equation \ref{eq:unary} measures the appearance evidence for landmark $i$ at location $l_i$. We write $f(l_i \mid {\bf I})$ for the HOG feature vector (see Section \ref{sec:impl_detail} for detail) extracted from $l_i$ in image ${\bf I}$. In the second term, the term $e(h(p_i),l_i \mid {\bf I})$ is equal to one minus the distance between $l_i$ and the closest segment boundary at segmentation level $h(p_i)$. This function penalizes landmarks being far from edges. The unary terms encourage locations with distinctive local appearances and with segment boundaries nearby to be identified as landmarks.
The binary term $\psi({\bf L},{\bf H} \mid {\bf I})$ is:
\begin{equation}
\psi({\bf L},{\bf H} \mid {\bf I}) = \sum_{(i,j) \in \mathcal{E}} {\bf w}_{i,j}^d \cdot d(l_i,l_j) + \sum_{\substack{(i,j) \in \mathcal{E}\\p_i=p_j}} {\bf w}_{i,j}^A \cdot A({\bf s}_{i,h},{\bf s}_{j,h} \mid {\bf I})\label{eq:binary}
\end{equation}
$d(l_i,l_j) = (-|x_i - x_j-\bar{x}_{ij}|,-|y_i - y_j-\bar{y}_{ij}|)$ measures the deformation cost for connected pairs of landmarks, where $\bar{x}_{ij}$ and $\bar{y}_{ij}$ are the anchor (mean) displacement of landmark $i$ and $j$. We adopt L1 norm to enhance our model's robustness to deformation. In the second term of Equation \ref{eq:binary}, $A({\bf s}_{i,h},{\bf s}_{j,h} \mid {\bf I}) = (\alpha(s_{i,h}^1,s_{j,h}^1 \mid {\bf I}),\alpha(s_{i,h}^1,s_{j,h}^2 \mid {\bf I}), \alpha(s_{i,h}^2,s_{j,h}^1 \mid {\bf I}),\alpha(s_{i,h}^2,s_{j,h}^2 \mid {\bf I}))$ is a vector storing the pairwise similarity between segments of nodes $i$ and $j$. This, together with the strength term ${\bf w}_{ij}^A$, models the SAC. The computation of $\alpha({\bf s}_{i,h},{\bf s}_{j,h} \mid {\bf I})$ is given in Section \ref{sec:impl_detail}. Finally, $\beta$ is a mixture-specific scalar bias.

The parameters of the score function are ${\bf \mathcal{W}}=\{{\bf w}^f_i\} \cup \{w^e_i\} \cup \{{\bf w}^d_{ij}\} \cup \{{\bf w}^A_{ij}\} \cup \{\beta\}$. Note that the score function is linear in ${\bf \mathcal{W}}$, therefore similar to \cite{2010_pami_felzenszwalb} we can express the model more simply by
\begin{equation}
S({\bf L}, {\bf H} \mid {\bf I}) = {\bf w} \cdot {\bf \varPhi}({\bf L}, {\bf H} \mid {\bf I})\label{eq:linear}
\end{equation}
where ${\bf w}$ is formed by concatenating the parameters ${\bf \mathcal{W}}$ into a vector.

\begin{figure}[t]
\centering
\includegraphics[width=\textwidth]{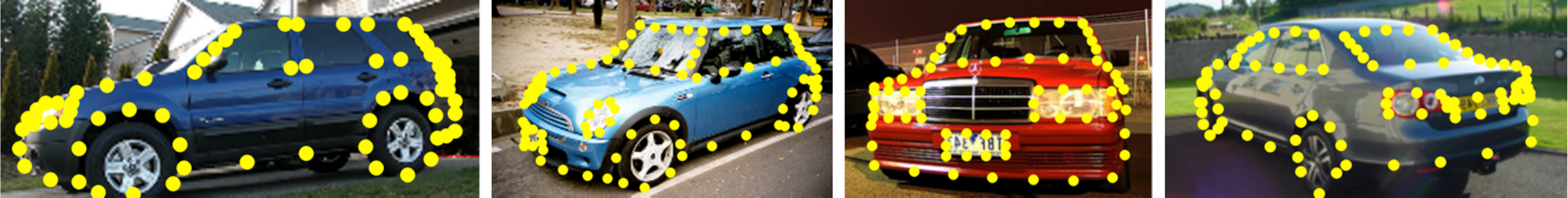}
\caption{The landmark annotations for typical images. Yellow dots are the chosen landmark locations. Please refer to Section \ref{sec:impl_detail} for landmark selection criteria.}
\label{fig:landmark}
\end{figure}

\subsection{Inference and Learning}

\noindent {\bf Inference.} The viewpoint $v$, the positions of the landmarks $\bf{L}$ and the segmentation levels ${\bf H}$ are unobserved. Our model detects the landmarks and searches for the optimal viewpoint and segmentation levels of parts simultaneously, as expressed by the following equation,
\begin{equation}
S({\bf I})=\max\limits_{v}[\max\limits_{{\bf H},{\bf L}} S({\bf L},{\bf H},v \mid {\bf I})]
\end{equation}
The outer maximizing is done by enumerating all mixtures. Within each mixture, we apply dynamic programming to estimate the segmentation levels and landmark positions of parts. Then the silhouette of each part can be directly inferred from its landmarks. In our experiment, it took a half to one minute to do the inference on an image about 300-pixel height. 

\noindent {\bf Learning.} We learn the model parameters by training our method for car detection (this is simpler than training it for part segmentation). We use a set of image windows as training data, where windows containing cars are labeled as positive examples and windows not containing cars are negative examples. A loss function is specified as:
\begin{equation}
\mathcal{J}({\bf w}) = \frac{1}{2}\|{\bf w}\|^{2}+C\sum\limits_{i}\max(0,1-t_{i}\cdot\max\limits_{{\bf L}_i,{\bf H}_i}{\bf w}\cdot {\bf \varPhi}({\bf L}_i,{\bf H}_i \mid {\bf I}_i))
\end{equation}
where $t_{i}\in\{1,-1\}$ is the class label of the object in the training image and $C$ is a constant. Let's take a closer look at the inner maximization. The segmentation levels of the semantic parts ${\bf H}$ are hidden and need to be estimated. The landmarks for the training images are not perfectly annotated (\eg they are not exactly on segment boundaries). To reduce the effect of such imprecision during the learning, we allow landmark locations to change within a small range (\ie the locations of landmarks become hidden variables), as long as shifted HOG boxes cover at least $60\%$ of the true HOG boxes. The CCCP algorithm \cite{yuille2003concave} is used to estimate the parameters by minimizing the loss function through alternating inference and optimization.

\begin{figure}[!t]
\centering
\includegraphics[width=0.8\textwidth]{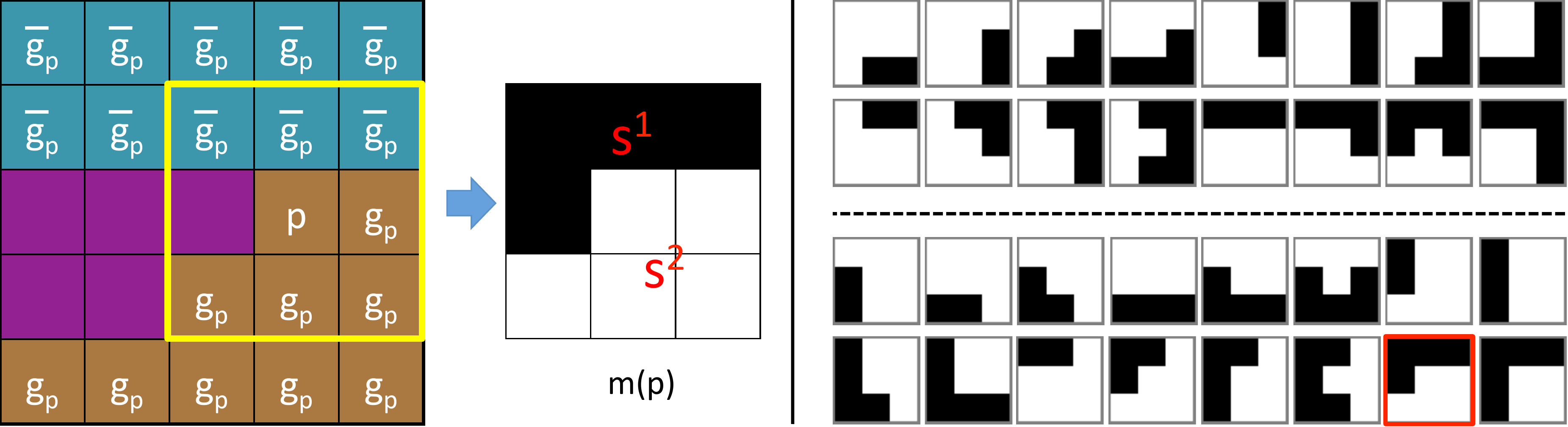}
\caption{Illustration of segment pair assignment. Right: The look-up table for segment pair assignment, which is divided into two parts (separated by the dashed line). White represents 1 and black represents 0. Left: an example of how to construct the binary matrix $m(p)$ for location $p$ and how to determine its segment pair. The hit of $m(p)$ in the look-up table is marked by the red rectangle. Best view in color.}\label{fig:seg_pair}
\end{figure}

\subsection{Implementation Details\label{sec:impl_detail}}

\noindent \textbf{Landmarks.} The landmarks are specified manually for each viewpoint. They are required to lie on the boundaries between the car and background (contour landmarks) or between parts (inner landmarks), so that the silhouettes of parts and the car itself can be identified from landmarks. 
For front/back view, we use 69 landmarks; for left and right side views, we use 74 landmarks; for the other views, we use 88 landmarks. The assignment of landmarks to parts is determined by the following rule: contour landmarks are assigned to parts they belong to (\eg landmarks of the lower half of wheels), and inner landmarks are assigned to parts that they are surrounding (\eg landmarks around license plates). See Figure \ref{fig:landmark} for some examples.

\noindent\textbf{Appearance features at landmarks}. The appearance features ${\bf f}$ at the landmarks are HOG features. More specifically, we calculate the HOG descriptor of an image patch centered at the landmark. The patch size is determined by the $80\%$ percentile of the distances between neighboring landmarks in training images.

\noindent\textbf{Appearance similarity between segments}. The similarity $\alpha(\cdot,\cdot)$ is a two dimensional vector, whose components are the $\chi^{2}$ distances of two types of features of the segments: color histograms and the grey-level co-occurrence matrices (GLCM) \cite{haralick1973textural}. The color histograms are computed in the HSV space. They have 96 bins, 12 bins in the hue plane and 8 bins in the saturation plane. The GLCM is computed as follows: We choose 8 measurements of the co-occurrence matrix, including HOM, ASM, MAX and means (variances and covariance) of x and y (please refer to \cite{haralick1973textural} for details); The GLCM feature is computed in the R, G and B channels in 4 directions (0, 45, 90, 135 degrees); As a result, the final feature length is 96 (8 measurements $\times$ 3 channels $\times$ 4 directions).

\noindent\textbf{Segment pair assignment.} To determine the segment pairs for locations on boundaries, we build a look-up table which consists of 32 3-by-3 binary matrices, as shown in the right of Figure \ref{fig:seg_pair}. At each boundary location $p$ we construct a 3-by-3 binary matrix $m(p)$ according to the segmentation pattern of its 3-by-3 neighborhood: locations covered by the same segment as $p$'s are given value 1 and other locations are given value 0. We denote the segment which $p$ belongs to by $g_p$, and the segment which most 0-valued locations in $m$ belong to by $\bar{g}_p$. See the left part of Figure \ref{fig:seg_pair} for an example. If $m(p)$ matches to one of the upper 16 matrices, $g_p$ will be $s^1$ and $\bar{g}_p$ be $s^2$ of $p$; If it matches to one of the lower 16 matrices, $\bar{g}_p$ will be $s^1$ and $g_p$ be $s^2$ of $p$. See the appendix for more information.  

\begin{figure*}[!t]
\centering
\subfigure[PASCAL VOC 2010]{\includegraphics[width=0.85\textwidth]{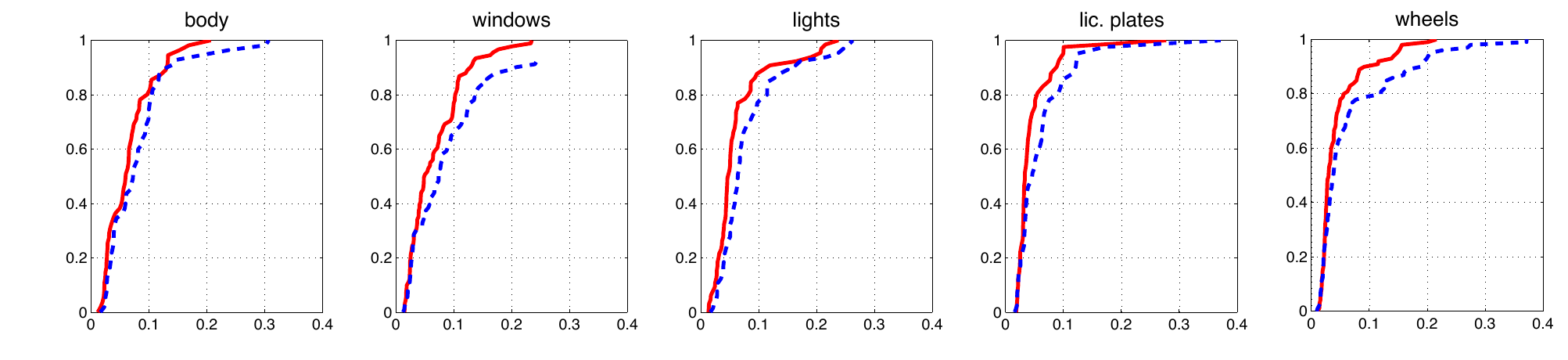}}
\subfigure[CAR3D]{\includegraphics[width=0.85\textwidth]{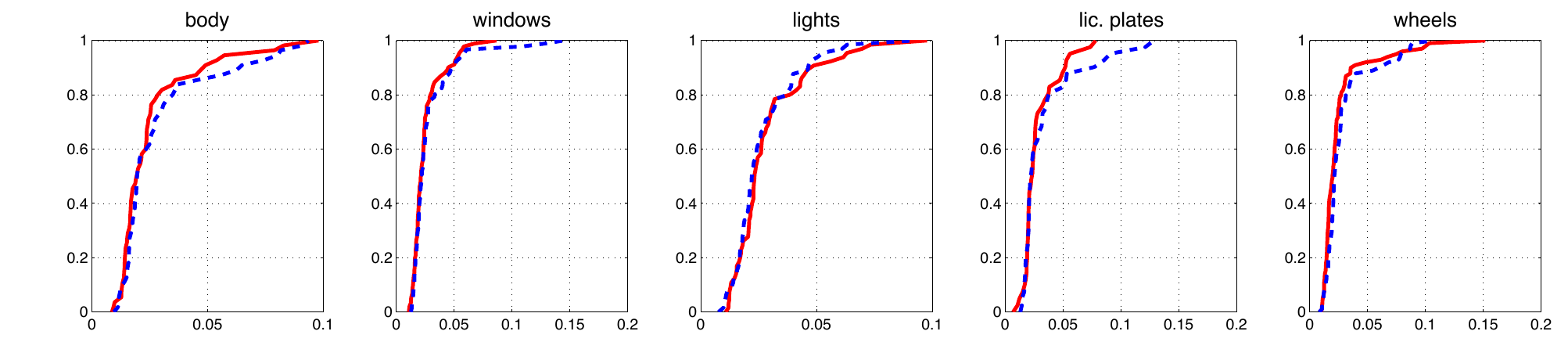}}
\caption{Cumulative localization error distribution for parts. X-axis is the average localization error normalized by image width, and Y-axis is the fraction of the number of testing images. The red solid lines are the performance using SAC and the blue dashed lines are the performance of \cite{2012_cvpr_zhu}.}
\label{fig:loc}
\end{figure*}

\begin{figure*}[t]
\centering
\subfigure[PASCAL VOC 2010]{\includegraphics[width=0.85\textwidth]{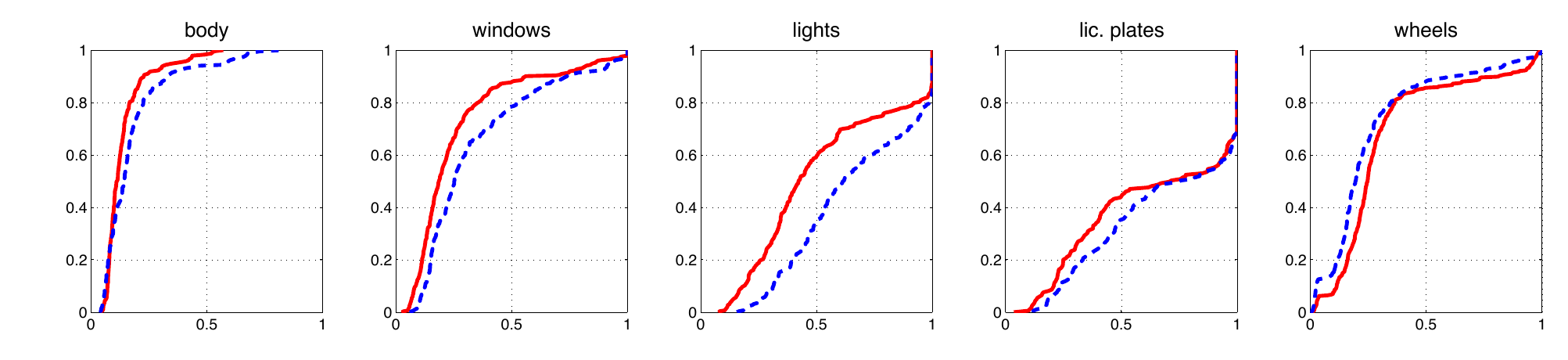}}
\subfigure[CAR3D]{\includegraphics[width=0.85\textwidth]{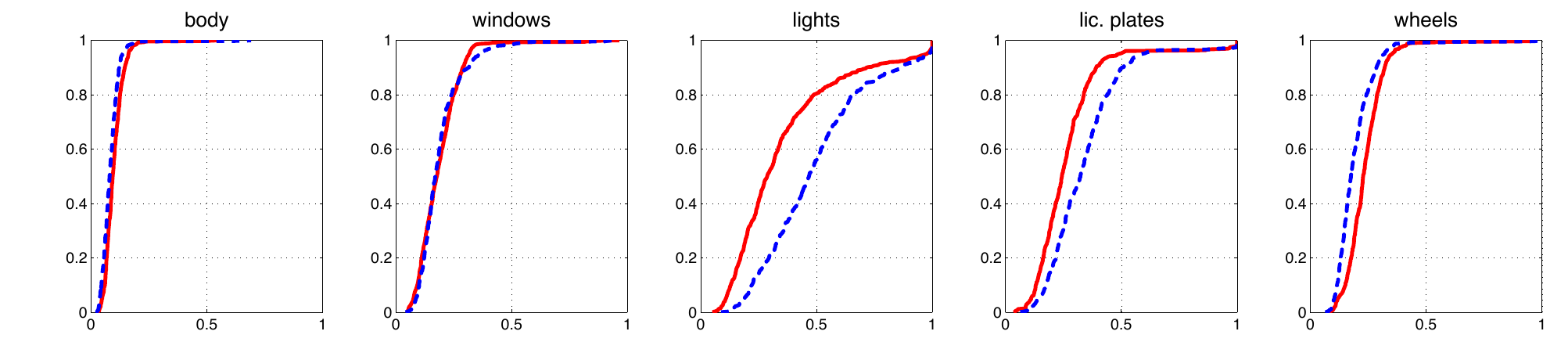}}
\caption{Cumulative segmentation error distribution for parts. X-axis is the average segmentation error normalized by image width, and Y-axis is the fraction of the number of testing images. The red solid lines are the performance using SAC and the blue dashed lines are the performance of \cite{2012_cvpr_zhu}.}
\label{fig:seg}
\end{figure*}

\begin{figure*}[t]
\centering
\includegraphics[width=0.85\textwidth]{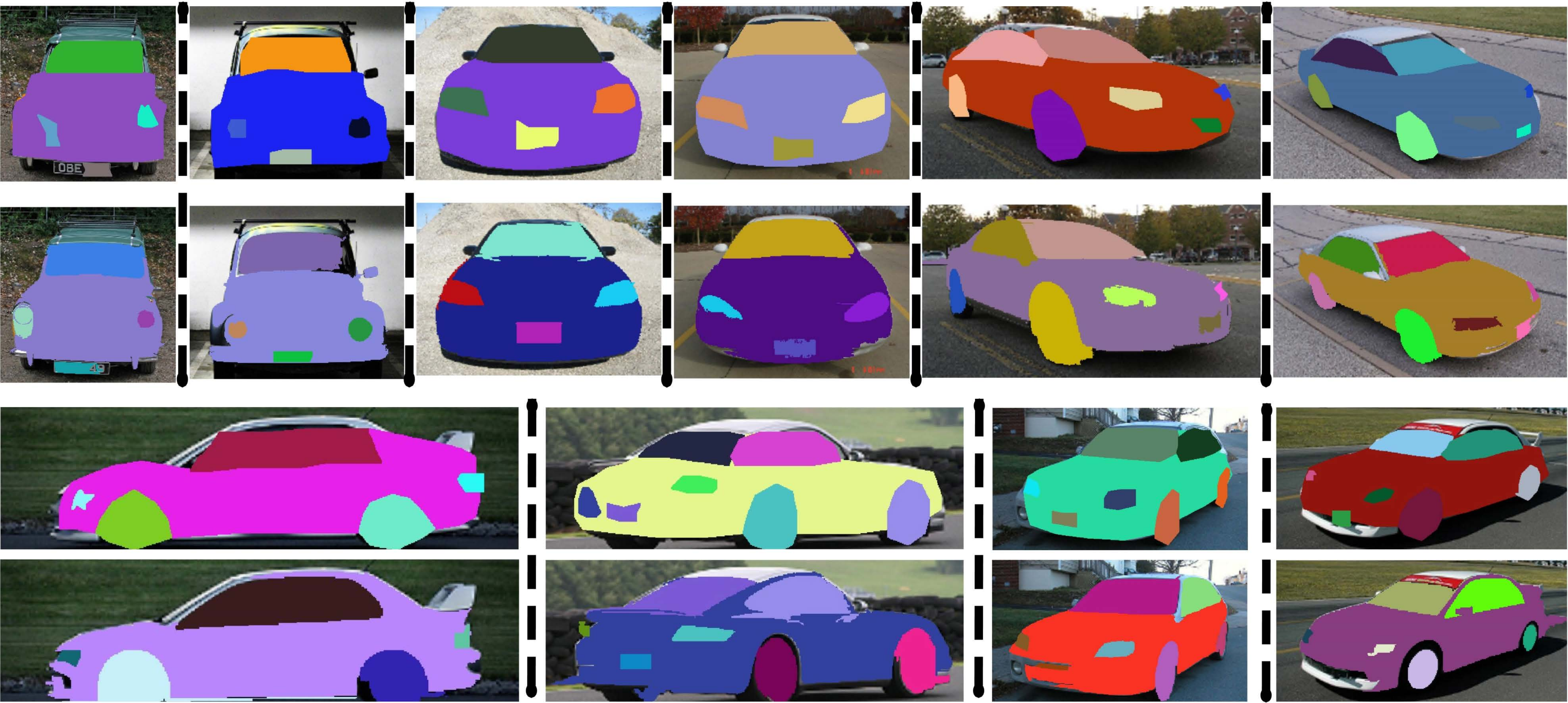}
\caption{Visualized comparison of our method with \cite{2012_cvpr_zhu} on car part segmentation. In each pair of results, the lower one is produced by our method.}\label{fig:compared}
\end{figure*}

\begin{figure*}[t]
\centering
\includegraphics[width=0.85\textwidth]{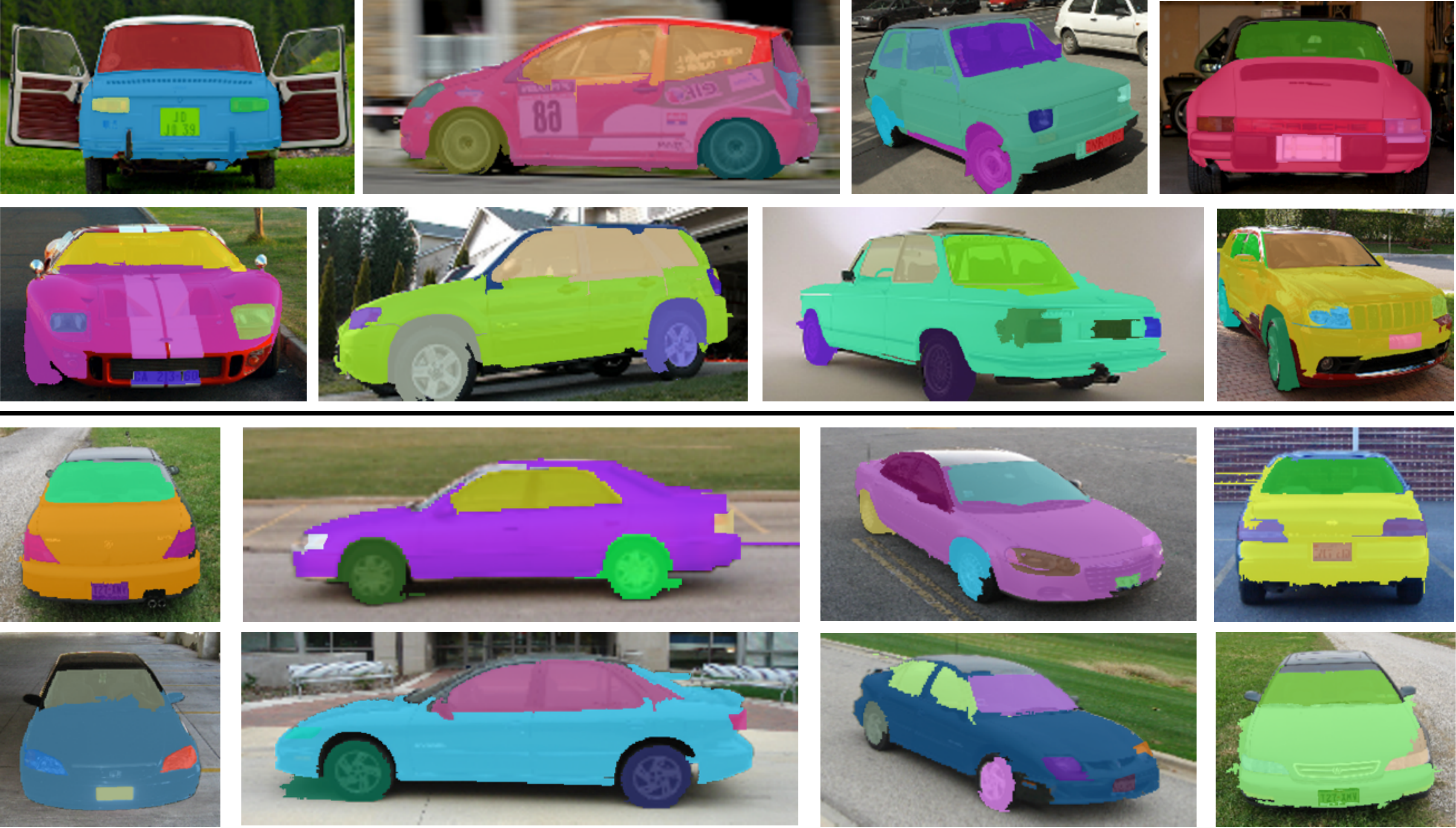}
\caption{More segmentation results of our method on VOC10 (upper) and CAR3D (lower).}\label{fig:more}
\end{figure*}

\section{Experiments \label{sec:experiments}}

\subsection{Dataset}

We validate our approach on two datasets, PASCAL VOC 2010 (VOC10) \cite{pascal-voc-2010} and 3D car (CAR3D) \cite{2007_iccv_savarese}. VOC10 is a hard dataset because the variations of the cars (e.g, appearance and shape) are very large. From VOC10, we choose car images whose sizes are greater than $80\times 80$. This ensures that the semantic parts are big enough for inference and learning. Currently our method cannot handle occlusion, so we remove images where cars are occluded by other objects or truncated by image boarder. We augment the image set by flipping the cars in the horizontal direction. This yields a dataset containing 508 cars. Then we divide images into seven viewpoints spanning over $180^\circ$ spacing at $30^\circ$. CAR3D provides 960 non-occluded cars. We also divide them into seven viewpoints (instead of using the original eight viewpoints). We collect 300 negatives images by randomly sampling from non-car images of PASCAL VOC 2010 using windows of the sizes of training images. These 300 negative images are used for both datasets. In our experiments, for each dataset, we randomly select half of the images as training data and test the trained model on the other half.

\subsection{Baseline}

We compare our method with the model proposed by Zhu and Ramanan \cite{2012_cvpr_zhu} on landmark localization and semantic part segmentation. We simply use their code to localize landmarks and assume the regions surrounded by certain landmarks are the semantic parts. Note that we use the same landmark and part definitions for both the baseline and our methods.

\subsection{Evaluation}
We first evaluate our method on landmark localization. We normalize the localization error as Zhu and Ramanan did in \cite{2012_cvpr_zhu}. In this and the following experiments, we consider parts of same category as a single part (\eg two lights of a front-view car are treated as one part). Figure \ref{fig:loc} shows the cumulative error distribution curves on both datasets.
We can see that by using SAC we had a big improvement of the landmark localization performance of all semantic parts on VOC10. We achieved better or comparable performance on CAR3D. Images in CAR3D are relatively easier than those in VOC10 and therefore SAC cannot bring big performance gain.

Then we evaluate our method on semantic part segmentation. The segmentation error of a part is computed by $(1-IOU)$, where $IOU$ is the intersection of detected segments and ground truth segments over the union of them. Figure \ref{fig:seg} shows the cumulative error distribution curves on both datasets. Again, using SAC our method improves the performance on almost all parts (improvement on lights and license plate is significant). However, we got slightly worse result on wheels. The errors occurred when SWA produces segments that are crossing the boundaries of wheels and the nearby background at all levels. The reason is that due to illumination and shading, it is difficult to separate wheels and background by appearance.

Figure \ref{fig:compared} shows the visualization comparison, from which we can see that our method works better on part boundaries, especially for lights and license plates. Figure \ref{fig:more} shows more segmentation results on VOC10 and CAR3D.

\section{Conclusion \label{sec:conclusion}}
In this paper, we address the novel task of car parsing, which includes obtaining the positions and the silhouettes of the semantic parts (e.g., windows, lights and license plates). We propose a novel graphical models which integrates the SAC coupling terms between neighboring landmarks, including using hidden variables to specify the segmentation level for each part. This allows us to exploit the appearance similarity of segments within different parts of the car. The experimental results on two datasets demonstrate the advances of using segment appearance cues. Currently, the model cannot handle large occlusion and truncation, which is our future direction.

\section{Acknowledgement}
This material is based upon work supported by the Center for Minds, Brains and Machines (CBMM), funded by NSF STC award CCF-1231216.

\begin{appendices}

\begin{figure}[t]
\centering
\subfigure[]{\includegraphics[width=0.4\textwidth]{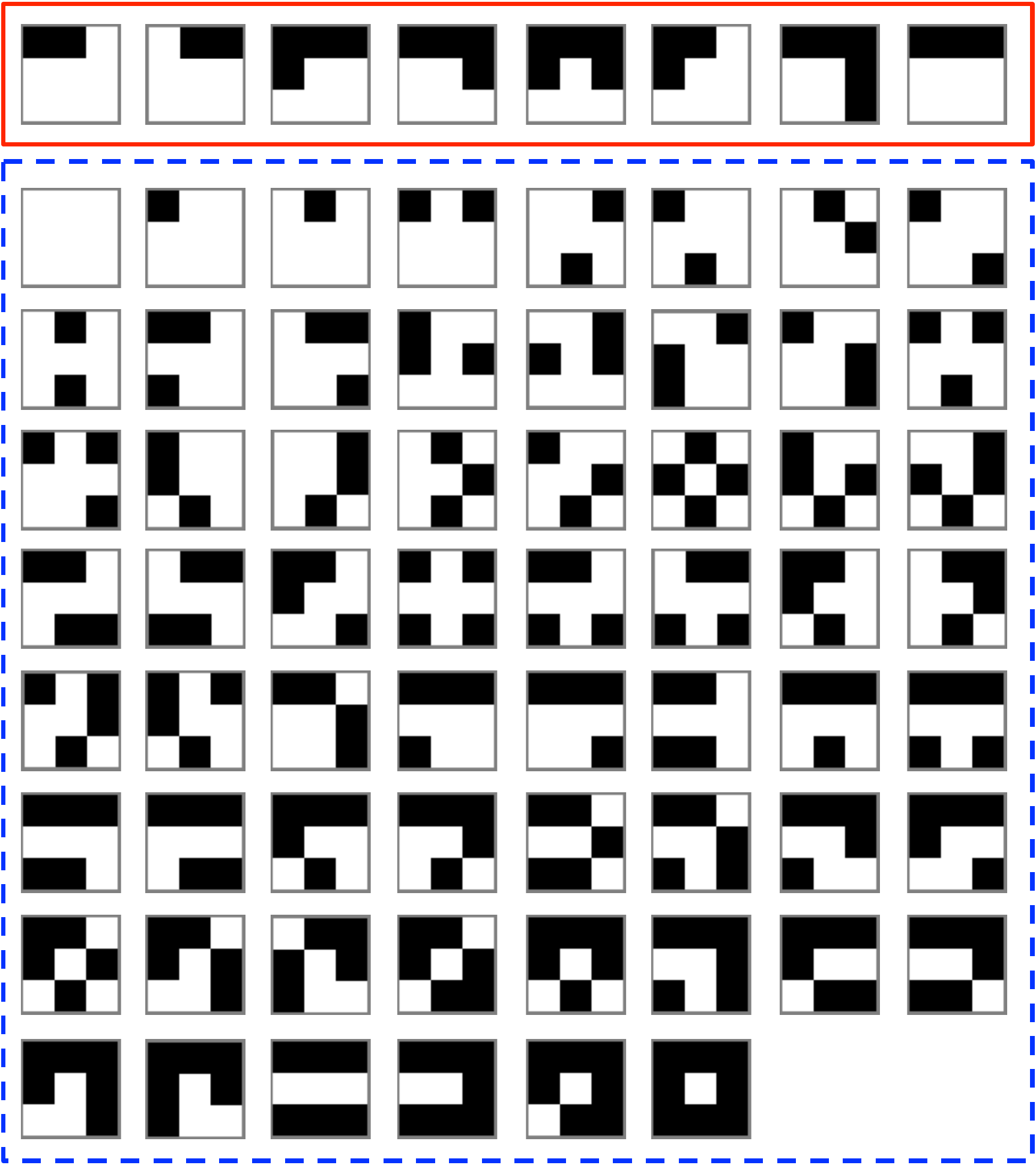}\label{fig:all}}
\subfigure[]{\includegraphics[width=0.4\textwidth]{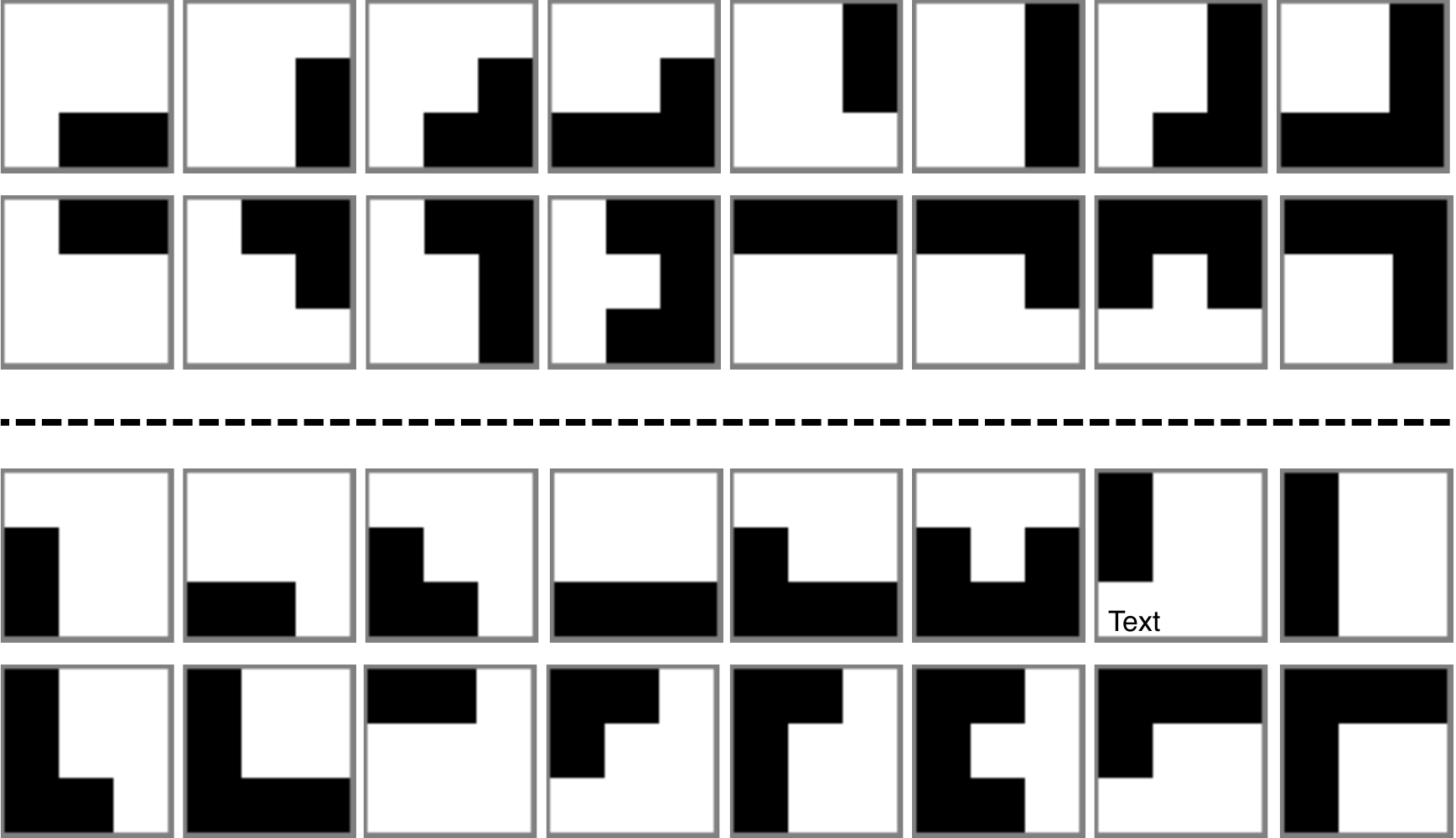}\label{fig:pattern}}
\caption{(a) 70 out of all 256 3-by-3 binary matrices (black indicates ``0" and white indicates ``1"), with the center fixed to one. Matrices in red rectangle are used to generated the 32 binary matrices of the look-up table. Matrices in the blue dashed rectangle are considered not suitable for indexing. (b) The 32 binary matrices in the look-up table, separated by a dashed line.}
\end{figure}

\section{Segment Pairs}

The look-up table is used to choose two segments from those near a boundary point and assign $s^1$ and $s^2$ to them. The first design criterion is that the assignment should be consistent, which is twofold: moving a contour point in its vicinity should not change its segment pair assignment; for two nodes in the graphical model whose landmarks are from the same part, their segment pairs should have the same order (e.g. both's $s^1$ are assigned to segments inside the part and $s^2$ assigned to segments outside the part) across different images. This criterion guarantees that learning the parameters (${\bf w}_{i,j}^A$ in equation 3) of SAC terms are statistically meaningful. The second criterion is that the look-up table should be able to identify locations with jagged edges, as in such locations it is very hard to guarantee consistency.

\subsection{Design of Look-Up Table}

We use 3-by-3 binary matrices to index the local segmentation patterns around contour locations. Figure \ref{fig:all} shows 70 out of all 256 possible matrices (the center is fixed to one), the rest of which are obtained by rotating these 70 prototypes. Not all of the 256 matrices are suitable for indexing. Some of them correspond to jagged edges and some of them will not occur on the contours in practice. We pick 8 matrices from the first row in figure \ref{fig:all} and rotate them to generate a set of 32 matrices which compose of the look-up table as shown in figure \ref{fig:pattern}.

\begin{figure}[t]
\centering
\includegraphics[width=0.48\textwidth]{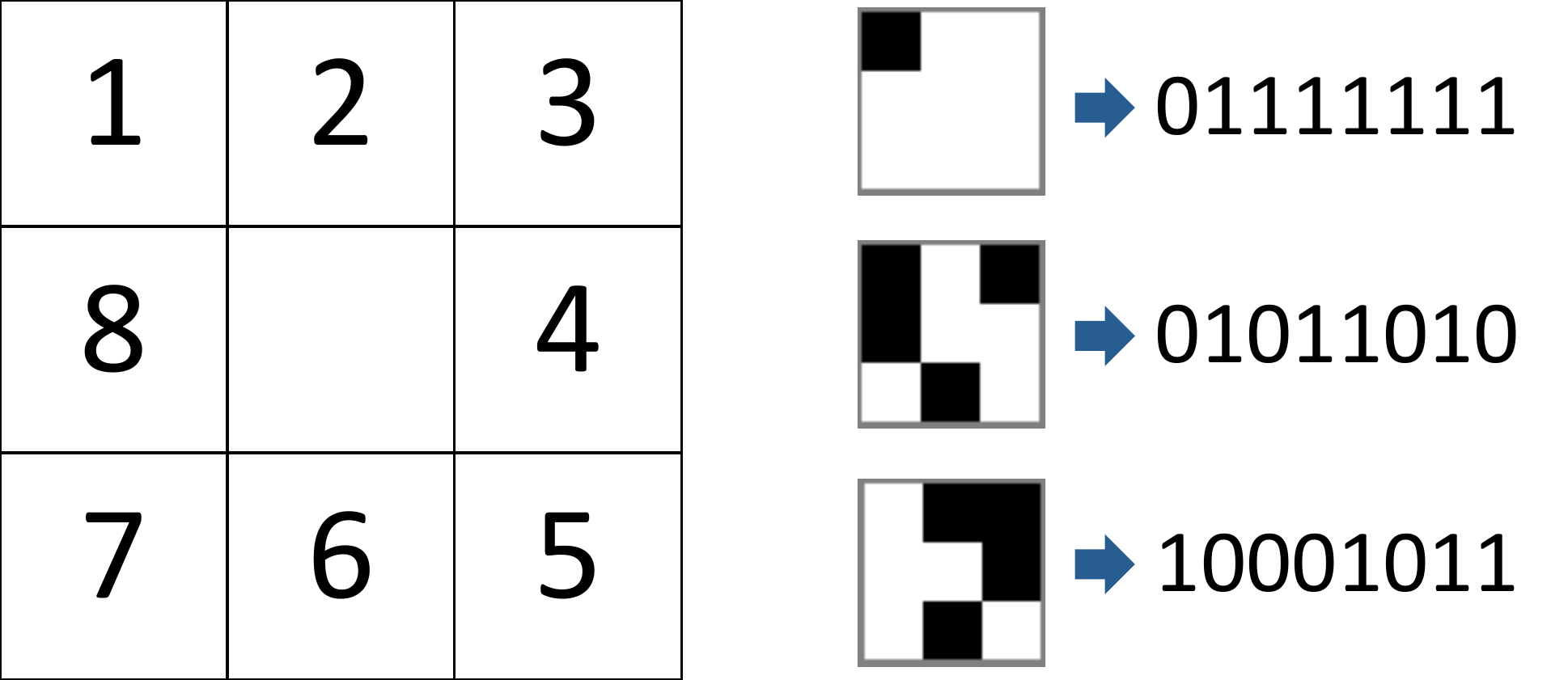}
\caption{Illustration of how to convert a binary matrix to a binary string. On the left, the numbers in the cells indicate the order of concatenation. On the right are three examples.}
\label{fig:order}
\end{figure}

More formally, for each binary matrix $m$, we convert it into an 8-bit binary string $b_m$ by concatenating its components in clockwise order starting from the upper left component. Then we use the matrices whose binary strings satisfy the following constraints
\begin{gather}
\sum_{i=1}^{8}\|b_m(i)-b_m(\mbox{mod}(i,8)+1)\| = 2\\
1 < \sum_{i=1}^8 (1-b_m(i)) < 6 
\end{gather}
where $b_m(i)$ is the i-th component of the binary string $b_m$. The first constraint says there should be exactly 2 jumps (i.e. ``1" to ``0" or ``0" to ``1") in $b_m$. The second one requires $b_m$ to have enough ``1". See figure \ref{fig:order} for some examples.

\begin{figure}[t]
\centering
\includegraphics[width=0.85\textwidth]{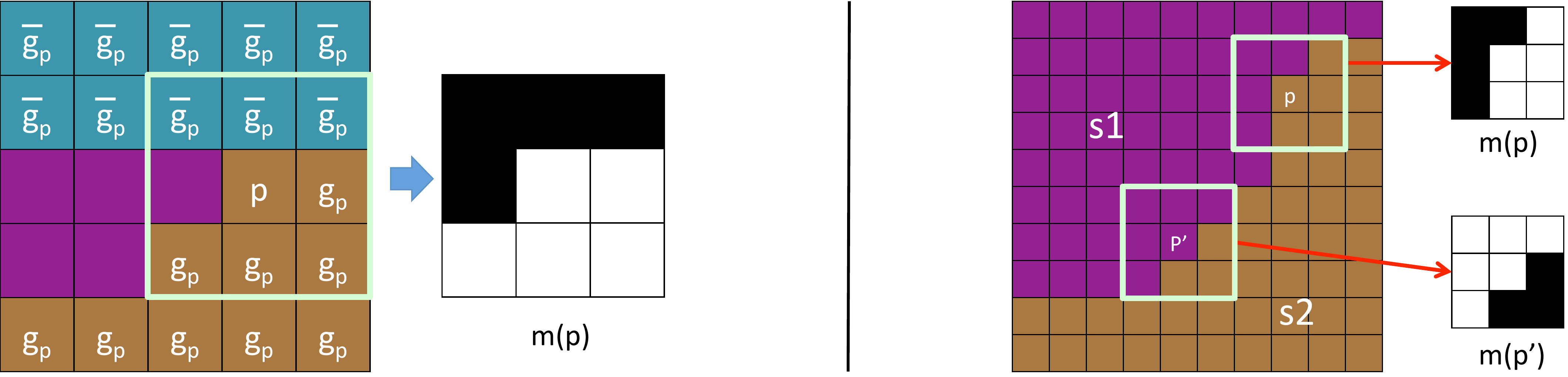}
\caption{Example of how segment pair assignment rule works (left) and illustration of its consistency (right).}
\label{fig:example}
\end{figure}

\subsection{Segment Pair Assignment on Contour Locations}

For convenience, we first repeat the rule of assigning $s^1$ and $s^2$ to two of the segments around a boundary location.  At each boundary location $p$ we construct a 3-by-3 binary matrix $m$ according to the segmentation pattern of its 3-by-3 neighborhood: locations covered by the same segment which covers $p$ are given value 1 and other locations are given value 0. We denote the segment which $p$ belongs to by $g_p$, and denote by $\bar{g}_p$ the segment which most $0$-valued locations in $m$ belong to. Then we search in the look-up table for the same binary matrix as $m$. If there is a hit from the upper 16 matrices in figure \ref{fig:pattern}, $g_p$ will be $s^1$ and $\bar{g}_p$ will be $s^2$; if there is a hit from the lower 16 matrices, $g_p$ will be $s^2$ of $p$ and $\bar{g}_p$ will be $s^1$ of $p$; otherwise, we will not apply SAC terms to $p$ in the score function.

The left of figure \ref{fig:example} shows how to compute the binary matrices for contour locations. The green rectangle marks the 3-by-3 neighborhood with $p$ in the center. The bronze segment is $g_p$ and the cyan segment is $\bar{g}_p$. According to the above rule, the bronze region is given value 1, and the rest is given value 0; then we got a hit in the look-up table, which assigns $s^2$ to the bronze segment and $s^1$ to the cyan segment.

On the right of figure \ref{fig:example}, we show two segmentation patterns from two locations $p$ and $p'$ not far from each other. Although different from each other, they all assign $s^1$ to the bronze segment and $s^2$ to the violet segment. In fact, in this example, all points along the segment boundaries have the same assignment (\ie bronze segment to $s^1$ and violet segment to $s^2$). This shows the consistency of the assignment algorithm.
\end{appendices}

\bibliographystyle{plain}
\bibliography{egbib}

\end{document}